\documentclass[letterpaper]{article} 
\usepackage[preprint]{aaai2027}  
\usepackage[hyphens]{url}  
\usepackage{graphicx} 
\usepackage{natbib}  
\usepackage{caption} 
\usepackage{algorithm}
\usepackage{algorithmic}
\usepackage{multirow}
\usepackage{amsmath,amssymb}

\usepackage{newfloat}
\usepackage{listings}
\DeclareCaptionStyle{ruled}{labelfont=normalfont,labelsep=colon,strut=off} 
\floatstyle{ruled}
\newfloat{listing}{tb}{lst}{}
\floatname{listing}{Listing}

\usepackage{booktabs}

\title{FUSE: Active Functional Affordance Grounding through Adaptive Semantic--Geometric Evidence Acquisition}
\author{
    Zhou Chen\textsuperscript{\rm 1}, Sathyanarayanan N. Aakur\textsuperscript{\rm 1}
}
\affiliations{
    \textsuperscript{\rm 1}CSSE Department\\
    
    Auburn University, Auburn, AL, 36849\\
    \{zzc0053,san0028\}@auburn.edu
}

\begin{document}

\maketitle

\begin{abstract}
Embodied agents must often identify and interact with objects based on their \emph{function} rather than their identity, requiring them to actively acquire observations that reveal discriminative functional evidence. Existing affordance grounding methods operate from fixed viewpoints and lack mechanisms for deciding where to look when functional cues are occluded or incomplete. We introduce \emph{Active Functional Affordance Grounding}, a new task in which an agent sequentially explores a scene to identify and spatially ground an object satisfying a functional query. To address this problem, we propose FUSE, an adaptive semantic--geometric evidence acquisition framework that combines explicit uncertainty-driven exploration with a learned amortized planner to efficiently select informative viewpoints. We further introduce a Habitat-based benchmark for evaluating active functional grounding. 
Experiments show that FUSE achieves the highest observed non-oracle
grounding performance while reducing computation by $1.33\times$
relative to fully explicit exploration, and remains effective across
multiple affordance knowledge sources. 
\end{abstract}


\section{Introduction}
\begin{figure}[t]
    \centering
    \includegraphics[width=\linewidth]{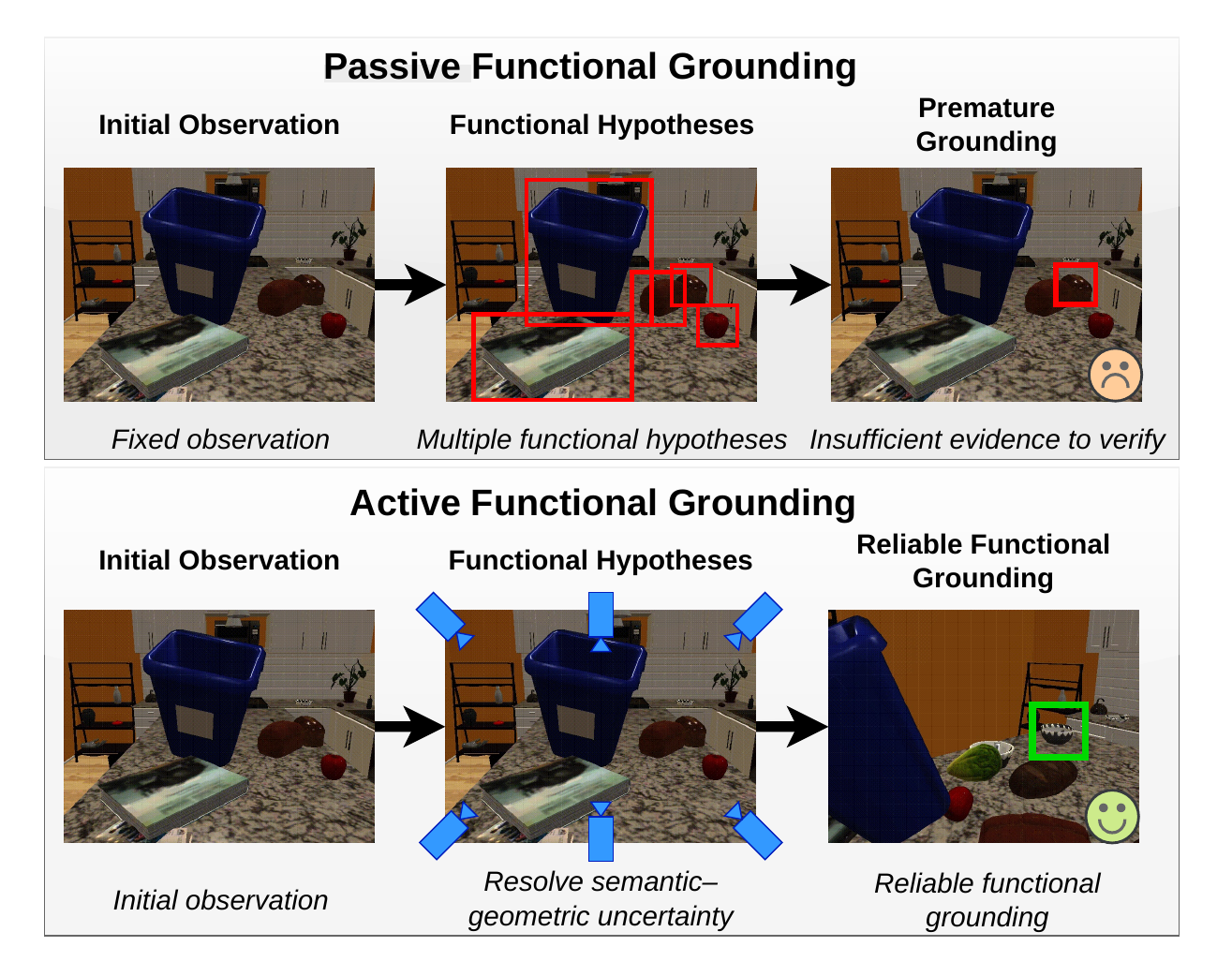}
    \caption{\textbf{Active functional affordance grounding.} Passive grounding predicts from a fixed observation and may commit before sufficient evidence is available. Active grounding acquires additional observations to reduce semantic–geometric uncertainty until a functionally suitable object is grounded.}
    \label{fig:intuition}
\end{figure}

Functional understanding of objects and their properties is central to embodied tasks such as long-horizon planning and problem solving~\cite{ahn2022saycan,10161317,zhu2023vima,pmlr-v229-zitkovich23a}. A functional concept is not fully captured by recognizing a familiar object~\cite{carion2025sam} or predicting a grasp from a fixed view~\cite{mahler2017dexnet2,Fang_2020_CVPR,9561877}. Instead, the same intent must support several coupled capabilities: inferring what properties a suitable object should possess, identifying candidate objects in a partially observed scene, selecting observations that resolve ambiguity, and producing a grounding that supports downstream interaction~\cite{chen2025craft,chen2025crafte,Nguyen-RSS-20,murali2020taskgrasp}. As illustrated in Fig.~\ref{fig:intuition}, a request such as ``cut a piece of rope'' or ``scoop food from a container'' may generate several plausible functional hypotheses, while the evidence needed to distinguish among them may be occluded, poorly localized, or unavailable from the initial viewpoint. We term this task \textit{active functional affordance grounding}: selecting sensing actions to determine and spatially ground which scene object satisfies a functional intent, without a predefined target class or identity. Candidate labels are only query-induced hypotheses, not search targets; the grounded object must be resolved from evidence acquired across viewpoints before physical execution and verification. 

\begin{figure*}
    \includegraphics[width=\linewidth]{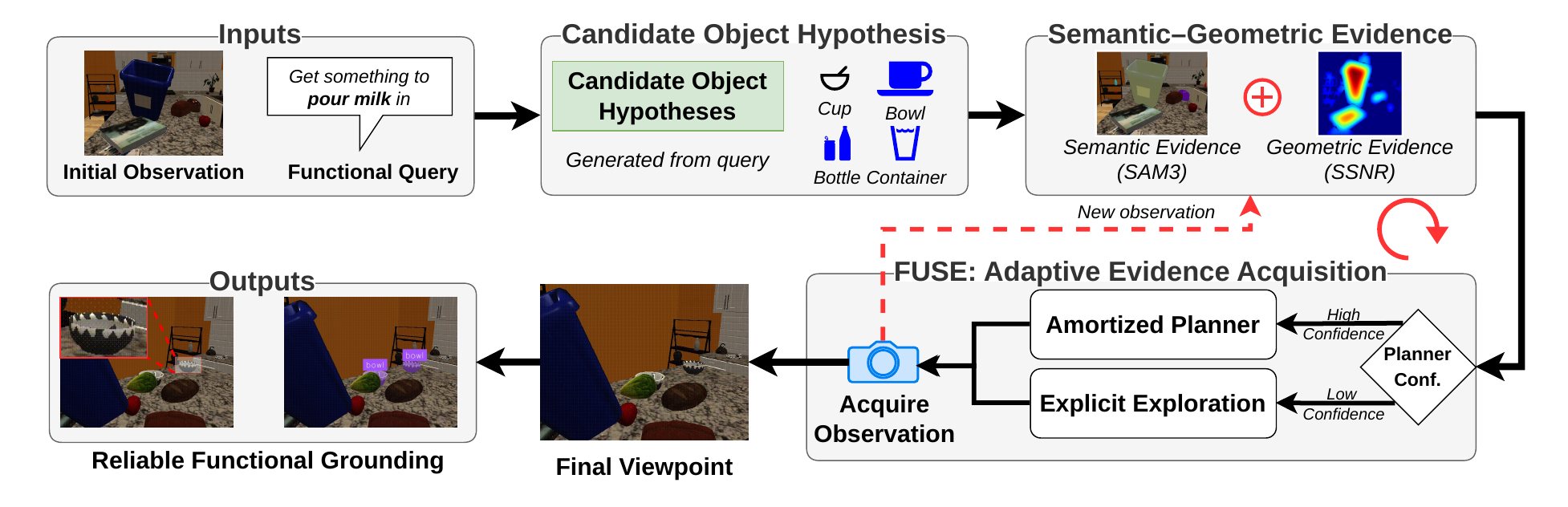}
    \caption{\textbf{Overview of FUSE.} A functional query is first converted into candidate object hypotheses and grounded using semantic (SAM3) and geometric (SSNR) evidence. FUSE adapts between fast amortized planning and explicit exploration to acquire informative observations until no further improvement is observed, returning the best reliable functional grounding.}
    \label{fig:framework}
\end{figure*}

This formulation places different requirements on perception than existing active or affordance-centered tasks. Passive affordance grounding predicts task-relevant objects, parts, or interaction regions from the available observation and may therefore commit before sufficient evidence has been acquired~\cite{chen2025craft,chen2025crafte,sawatzky2019object,li2022toist}. Active recognition and object search acquire additional observations, but typically seek a predefined category or instance~\cite{Lei_2026_CVPR,chaplot2020objectnav,batra2020objectnav}. Task-oriented view planning can improve visibility or grasp confidence, yet generally begins with an already selected target object or manipulation hypothesis~\cite{shi2025viso,tong2026gcngrasp}. In contrast, active functional affordance grounding (illustrated in Fig.~\ref{fig:intuition}) requires the complete sensing process: forming candidate hypotheses, identifying which semantic and geometric evidence remains unresolved, and acquiring observations until a suitable target can be reliably grounded.

To address this challenge, we introduce FUSE (\textit{Functional Understanding through Semantic--Geometric Exploration}), an adaptive evidence-acquisition framework for active functional affordance grounding. FUSE is built on the principle that sensing should reduce the unresolved semantic and geometric evidence needed to ground a functional target, which we refer to as \textit{functional uncertainty}. As illustrated in Fig.~\ref{fig:framework}, FUSE combines two complementary planning modes: (1) an amortized planner that rapidly predicts which sensing action is most likely to reduce functional uncertainty, and (2) an explicit planner that evaluates candidate actions using semantic and geometric predictions from an incrementally refined spatial scene representation. When the amortized predictions are confident, FUSE avoids costly search and representation updates; when candidate actions are ambiguous or the task evidence remains unresolved, it invokes explicit planning. By adaptively switching between fast prediction and deliberate search, FUSE preserves reliable functional grounding while reducing the computational cost of active perception.

The \textbf{contributions} of this paper are four-fold: (i) we introduce, define, and formulate \emph{active functional affordance grounding} as a new sequential perception task in which a functional query must guide target discovery and spatial grounding under partial observability; (ii) we develop and will release a Habitat-based benchmark and evaluation protocol that operationalizes the task through functional queries, controlled sensing trajectories, standardized stopping and success criteria, sensing budgets, oracle controls, and evaluation tracks spanning static, canonical-view, VLM-active, search-based, amortized, and adaptive methods; (iii) we present FUSE, a confidence-gated adaptive evidence-acquisition framework that alternates between fast amortized planning and explicit semantic--geometric search over an incrementally refined spatial scene representation; and (iv) through controlled evaluations, we show that task-conditioned semantic–geometric evidence acquisition improves active grounding across knowledge sources.

\begin{figure*}[t]
    \centering
    \begin{tabular}{cccc}
        \includegraphics[width=0.45\columnwidth]{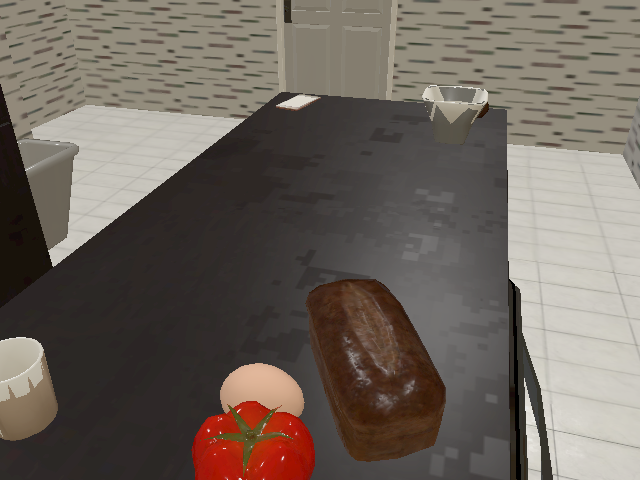} &
        \includegraphics[width=0.45\columnwidth]{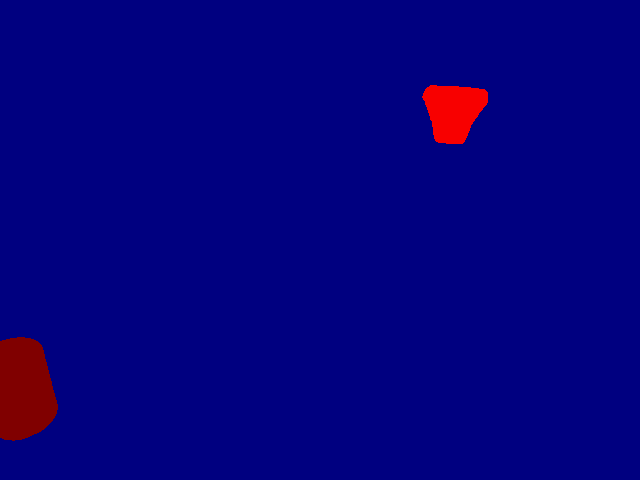} &
        \includegraphics[width=0.45\columnwidth]{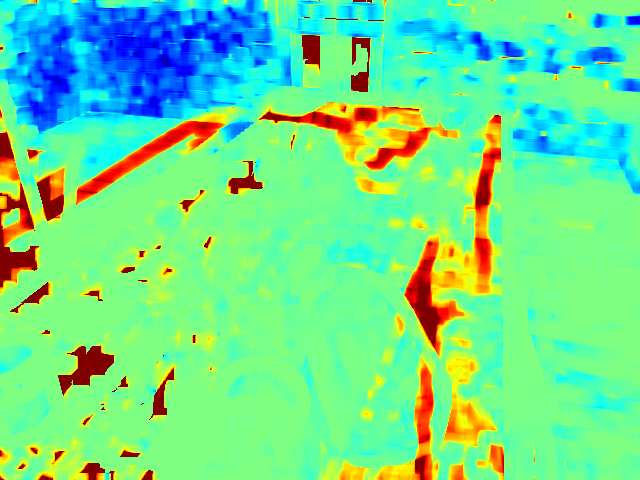} &
        \includegraphics[width=0.45\columnwidth]{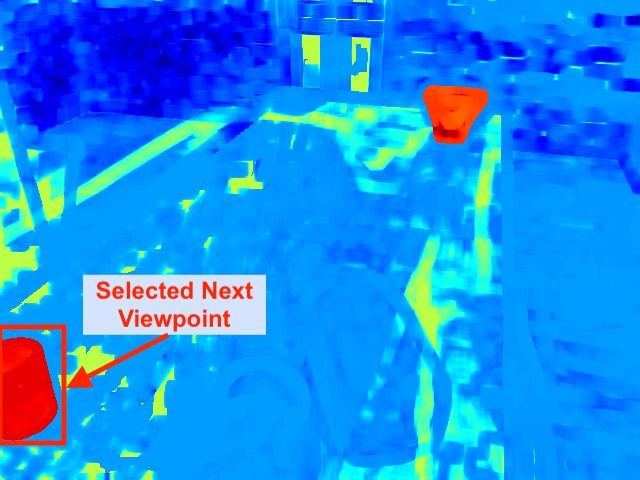} \\
        \small (a) Input Observation &
        \small (b) Semantic Evidence & 
        \small (c) Geometric Uncertainty &
        \small (d) Fused Evidence Map
    \end{tabular}
    \caption{
    Visualization of FUSE's semantic--geometric evidence acquisition. SAM3 highlights candidate functional objects, SSNR estimates geometric uncertainty, and their fusion produces the evidence map used to select the next viewpoint. 
    }
    \label{fig:evidence_maps}
\end{figure*}

\section{Related Work}
\textbf{Functional Affordance Grounding and Task-Oriented Manipulation.}
Recent work has increasingly focused on grounding functional language queries to objects, object parts, and grasp configurations for embodied manipulation. Neuro-symbolic affordance grounding methods such as CRAFT and CRAFT-E~\cite{chen2025craft,chen2025crafte} combine visual perception with structured functional reasoning to identify objects satisfying task-level affordance queries from a single observation. Foundation-model-based approaches extend this capability using large vision--language models and multimodal reasoning~\cite{yu2025seqafford,ahn2022saycan}, while task-oriented grasping methods leverage affordance prediction to optimize grasp synthesis or next-best-view selection for manipulation~\cite{zhang2023affordance,tong2026gcngrasp,Lei_2026_CVPR,shi2025viso}. These approaches have substantially improved functional reasoning from available observations; however, they generally assume that sufficient visual evidence has already been observed. In contrast, Active Functional Affordance Grounding explicitly addresses the preceding perception problem: determining \emph{what additional observations should be acquired} before reliable functional grounding. 

\textbf{Active Recognition, Search, and Next-Best-View Perception.}
Active perception methods improve recognition, localization, and scene understanding by selecting informative sensing actions. Prior work has explored active object recognition~\cite{paletta2000active}, object search and navigation~\cite{chaplot2020objectnav}, next-best-view planning for reconstruction and inspection~\cite{strong2024nextbestsense}, and active grasping systems that reposition the camera to improve grasp estimation~\cite{Lei_2026_CVPR,shi2025viso}. More recently, spatial world models and 3D Gaussian Splatting have enabled active exploration through explicit scene representations~\cite{kerbl2024hierarchical,chen2025ease}. These methods optimize sensing for a known recognition, localization, reconstruction, or manipulation objective, where the target is specified a priori or the objective is generic scene coverage. In contrast, Active Functional Affordance Grounding must first infer candidate functional targets from a language query and then actively acquire observations until sufficient evidence has been accumulated for grounding.

\textbf{Affordance-Guided Active Grasping and Adaptive Planning.}
Several works actively reposition the camera to improve grasp synthesis, manipulation success, or affordance estimation~\cite{Lei_2026_CVPR,shi2025viso,tong2026gcngrasp}, while recent embodied systems combine learned policies with explicit planning or world models to balance efficiency and accuracy during decision making~\cite{chen2025ease,assran2023self,ha2018world}. These approaches typically assume that the target object has already been identified and focus on improving its perception or selecting an appropriate manipulation strategy. In contrast, FUSE performs adaptive evidence acquisition before manipulation, using a confidence-gated combination of amortized planning and explicit semantic--geometric reasoning to determine \emph{which} candidate object satisfies the functional query and to continue exploration until grounded.

\section{Proposed Approach}

\noindent\textbf{Problem Formulation.}
Given a functional query ($q$), an affordance-generation module first produces a set of candidate object hypotheses $\mathcal{O}_q{=}\{o_1,\ldots,o_N\}$ that may satisfy the requested function. Given an initially partial observation ($x_0$) and a set of feasible sensing actions ($\mathcal{A}$), the goal of active functional affordance grounding is to identify and spatially ground an object $o^* \in \mathcal{O}_q$ without being given its identity in advance. At each step $t$, the agent selects an action $a_t \in \mathcal{A}$, acquires a new observation $x_{t+1}$, and integrates the resulting evidence into its current scene state $\mathcal{B}_{t+1}$. This evidence may capture semantic compatibility, object localization, visibility, and geometric structure. The process terminates when a candidate object $o^*$ can be grounded with sufficient confidence under the task-specific grounding criterion $\mathcal{V}(o^*,\mathcal{O}_q,\mathcal{B}_T)$. 
We formulate sensing as minimizing unresolved functional uncertainty while accounting for the cost of active perception:
\begin{equation}
\min_{T,\,a_{1:T},\,o^*}
\quad
\mathcal{U}_{\mathrm{func}}(\mathcal{O}_q,\mathcal{B}_T)
+
\lambda \sum_{t=1}^{T} C(a_t,\mathcal{B}_t),
\end{equation}
subject to $\mathcal{V}(o^*,\mathcal{O}_q,\mathcal{B}_T) \geq \tau$. Here, $\mathcal{U}_{\mathrm{func}}$ represents the unresolved semantic and geometric evidence required to ground the functional hypotheses induced by $q$, $C$ captures sensing and scene-representation update costs, and $\tau$ defines the required grounding confidence. Unlike next-best-view planning, the objective is not to reconstruct the entire scene, but to acquire evidence for reliable functional grounding.

\noindent\textbf{Functional Uncertainty}
In our framework, \textit{functional uncertainty} is represented as a spatial field conditioned on the candidate object hypotheses $\mathcal{O}_q$. Rather than assigning uncertainty to a predefined target object, the field identifies image regions in which the evidence required for grounding the functionally relevant candidates remains unresolved. This scene-centric formulation is appropriate because the target identity is not known in advance and must emerge through active evidence acquisition. At time $t$, we decompose the functional uncertainty into semantic-geometric components:

\begin{equation}
\mathbf{U}^{\mathrm{func}}_t(\mathcal{O}_q) = \mathbf{U}^{\mathrm{sem}}_t(\mathcal{O}_q)
+
\mathbf{U}^{\mathrm{geo}}_t.
\end{equation}

The semantic component captures where the current observation provides insufficient grounding evidence for the candidate hypotheses. The object hypotheses $\mathcal{O}_q$ are supplied to SAM3~\cite{carion2025sam}. For each candidate $o_i \in \mathcal{O}_q$, SAM3 predicts a mask $m_{t,i}$ with score $s_{t,i}$. The corresponding score is assigned to all pixels within the mask, and the candidate maps are aggregated to produce a dense semantic evidence map $\mathbf{S}_t(\mathcal{O}_q)$. After normalization to $[0,1]$, semantic uncertainty is estimated as
$\mathbf{U}^{\mathrm{sem}}_t(\mathcal{O}_q)=1-\mathbf{S}_t(\mathcal{O}_q)$.

The geometric component captures regions that are poorly explained by the current spatial scene representation. Given the observed image $x_t$ and the corresponding rendered image $\hat{x}_t$, we compute a dense structural signal-to-noise ratio (SSNR) map from discrepancies between their image gradients. High SSNR indicates strong agreement between the observation and rendering, whereas low SSNR indicates incomplete or inaccurately modeled structure. We invert and normalize this map to obtain $\mathbf{U}^{\mathrm{geo}}_t$, assigning larger uncertainty to regions with weaker reconstruction agreement. The resulting map jointly localizes where candidate-relevant semantic evidence is weak and where additional geometric observation may be useful. 
Figure~\ref{fig:evidence_maps} illustrates this for a functional query ``sip'' where the target object is a cup. 
A sensing action is considered informative when the consequent observation is expected to reduce this uncertainty while strengthening the evidence needed for reliable functional grounding.

\noindent\textbf{Spatial Evidence Representation.} 
We instantiate the scene state $\mathcal{B}_t$ as a persistent 3D spatial evidence representation that accumulates observations acquired during exploration. Reliable functional grounding requires evidence to be integrated across viewpoints in a common coordinate frame, rather than evaluated independently or retained only as an observation sequence. This spatial organization enables the agent to relate task-relevant evidence across camera poses, reason about visibility and occlusion, predict observations from candidate viewpoints, and localize where geometric evidence remains incomplete. Its purpose is not complete scene reconstruction; instead, it retains the spatial information needed to support functional grounding and subsequent sensing decisions. We implement this representation using an incrementally refined 3D Gaussian Splatting (3DGS) model~\cite{kerbl3Dgaussians}. The representation is initialized from a small set of seed observations registered through structure-from-motion.

At time $t$, the current scene representation model ($\mathcal{G}_t$) encodes the scene using a set of 3D Gaussians with position, appearance, scale, orientation, and opacity parameters. Given a camera pose $p_t$, the renderer produces a predicted observation $\hat{x}_t=\mathcal{R}(\mathcal{G}_t,p_t)$, where $\mathcal{R}$ denotes differentiable Gaussian rendering. Comparing the predicted image $\hat{x}_t$ with the acquired observation $x_t$ yields the SSNR-based geometric uncertainty used in the functional uncertainty map. After executing a sensing action and observing $x_{t+1}$, the new image is registered to the current representation and may be incorporated through incremental 3DGS refinement:
\begin{equation}
\mathcal{D}_{t+1} = \mathcal{D}_t\cup\{(x_{t+1},p_{t+1})\}; 
\mathcal{G}_{t+1} = \operatorname{Update}(\mathcal{G}_t,\mathcal{D}_{t+1}),
\label{eqn:3dgs_update}
\end{equation}
where $\mathcal{D}_{t+1}$ denotes all observations registered thus far. 
The updated representation improves predictions from nearby viewpoints and reduces uncertainty in newly observed regions. It provides the explicit spatial substrate needed to evaluate candidate viewpoints, compare predicted and observed evidence, and support subsequent sensing decisions.

\subsubsection{Functional Evidence Exploration.} 
Explicit evidence exploration uses the semantic and geometric components of functional uncertainty to determine where the agent should observe next. At time $t$, the SAM3 semantic evidence map $\mathbf{S}_t(\mathcal{O}_q)$ identifies regions likely to contain the candidate objects associated with the functional query, while the SSNR-based geometric uncertainty map $\mathbf{U}^{\mathrm{geo}}_t$ identifies regions that remain poorly explained by the current spatial evidence representation $\mathcal{G}_t$. 
Although functional uncertainty is defined using low semantic confidence, action selection uses positive semantic evidence to restrict exploration to candidate-relevant regions, while geometric uncertainty favors regions whose structure remains under-observed. 
We combine these maps to obtain a functional evidence-acquisition map:
\begin{equation}
\mathbf{E}_t(\mathcal{O}_q) = \mathbf{S}_t(\mathcal{O}_q) +
1-\operatorname{Norm}\!\left(\frac{\|\nabla x_t\|^2} {\|\nabla x_t-\nabla\hat{x}_t\|^2+\epsilon} \right)
\label{eq:evidence_map}
\end{equation}
where the gradient ratio is computed locally at each pixel and
normalized spatially to $[0,1]$ using $\operatorname{Norm}(\cdot)$,
such that larger values of the inverted term indicate greater geometric uncertainty.
Hence, Equation~\ref{eq:evidence_map} acts as an acquisition surrogate that balances candidate relevance with geometric uncertainty to explore regions that are both semantically promising and geometrically under-observed. 
For each feasible sensing action $a \in \mathcal{A}_t$, we define an associated image region $\Omega(a)$ that the corresponding camera motion is expected to reveal more clearly. 
Horizontal actions use the corresponding left or right image half, while tilt actions use the upper or lower half, respectively. 
We evaluate each action by averaging the evidence-acquisition map over its associated region: 

\begin{equation}
Q_t(a;\mathcal{O}_q) = \frac{1}{|\Omega(a)|}
\sum_{u \in \Omega(a)}
\mathbf{E}_t(\mathcal{O}_q)(u).
\end{equation}

The explicit exploration policy then selects
$a_t^*=\arg\max_{a \in \mathcal{A}_t} Q_t(a;\mathcal{O}_q)$.
This produces task-directed sensing: SAM3 evidence guides the agent toward regions associated with functionally relevant object hypotheses, while SSNR uncertainty encourages views that improve incomplete or poorly modeled scene structure. 
After executing $a_t^*$, the agent acquires a new observation $x_{t+1}$ and registers it to the spatial evidence representation. $\mathcal{G}_t$ is refined using the new view (Eqn.~\ref{eqn:3dgs_update}), after which the predicted observation, SSNR map, SAM3 evidence map, and evidence-acquisition map are recomputed. Updating the representation at every step is necessary for explicit exploration because the geometric uncertainty estimate must reflect all observations accumulated so far. Otherwise, regions that have already been sufficiently observed may remain uncertain and attract the policy.

\begin{algorithm}[t]
\caption{FUSE: Adaptive Evidence Acquisition}
\label{alg:fuse}
\begin{algorithmic}[1]
\REQUIRE Query $q$, action set $\mathcal{A}$, entropy threshold $\eta$, patience $K$, maximum steps $T$
\ENSURE Best observed grounding, candidate label, and viewpoint

\STATE Generate candidate object hypotheses $\mathcal{O}_q$ from $q$
\STATE Initialize $\mathcal{G}_0$ and registered seed-view set
$\mathcal{D}_0$
\STATE $r_{\mathrm{best}}\leftarrow-\infty$, $k_{\mathrm{fail}}\leftarrow 0$

\FOR{$t=0$ to $T-1$}
    \STATE Observe $x_t$ at viewpoint $v_t$
    \STATE Register $x_t$ and $\mathcal{D}_t {\leftarrow} \mathcal{D}_{t-1}{\cup}\{(x_t,v_t)\}$
    \STATE Obtain SAM3 masks and scores
    $\{(m_{t,i},s_{t,i}){:}o_i{\in}\mathcal{O}_q\}$
    \STATE Compute depth scores $d_{t,i}$ over each mask $m_{t,i}$
    \STATE $r_{t,i}\leftarrow s_{t,i}+d_{t,i}$
    \STATE $r_t \leftarrow \max_{o_i\in\mathcal{O}_q}r_{t,i}$ and
    $o_t \leftarrow \arg\max_{o_i\in\mathcal{O}_q}r_{t,i}$

    \IF{$r_t > r_{\mathrm{best}}$}
        \STATE $r_{\mathrm{best}} \leftarrow r_t$
        \STATE $o_{\mathrm{best}} \leftarrow o_t$, 
        $v_{\mathrm{best}} \leftarrow v_t$
        \STATE $k_{\mathrm{fail}} \leftarrow 0$
    \ELSE
        \STATE $k_{\mathrm{fail}} \leftarrow k_{\mathrm{fail}} + 1$
    \ENDIF

    \STATE \textbf{if} $k_{\mathrm{fail}} \geq K$ \textbf{then break}

    \STATE Select object context $\tilde{o}_t \in \mathcal{O}_q$
    for amortized planning
    \STATE Predict values $\hat{Q}_t(a;\tilde{o}_t)$ for
    $a\in\mathcal{A}_t$ and compute action entropy $H_t$

    \IF{$H_t \leq \eta$}
        \STATE $a_t \leftarrow
        \arg\max_{a\in\mathcal{A}_t}\hat{Q}_t(a;\tilde{o}_t)$
    \ELSE
        \STATE Refine $\mathcal{G}_t$ jointly over $\mathcal{D}_t$
        \STATE Compute $\mathbf{E}_t(\mathcal{O}_q)$ and explicit
        scores $Q_t(a;\mathcal{O}_q)$
        \STATE $a_t \leftarrow
        \arg\max_{a\in\mathcal{A}_t}Q_t(a;\mathcal{O}_q)$
    \ENDIF

    \STATE Execute $a_t$
\ENDFOR

\STATE \textbf{return} grounding of $o_{\mathrm{best}}$
at $v_{\mathrm{best}}$
\end{algorithmic}
\end{algorithm}

\begin{table*}[t]
\centering
\resizebox{\textwidth}{!}{
\begin{tabular}{lccccccc}
\toprule
\textbf{Method}
& \textbf{Success Rate} $\uparrow$
& \textbf{Avg. IoU} $\uparrow$
& \textbf{Avg. Steps} $\downarrow$
& \textbf{Target Pixel Ratio} $\uparrow$
& \textbf{Center Score} $\uparrow$
& \textbf{Distance Score} $\uparrow$
& \textbf{Final SAM3 Score} $\uparrow$ \\
\midrule

\multicolumn{8}{c}{\textit{Passive and Semi-Static Grounding}} \\
\midrule
Static
& 42.00 & 41.25 & 0.00 & 17.21 & 40.85 & 24.11 & 48.20\% \\
Canonical View
& 32.00 & 31.36 & 1.00 & 22.83 & 29.91 & 51.13 & 28.22\% \\

\midrule
\multicolumn{8}{c}{\textit{Active Baselines}} \\
\midrule
Random Active
& 56.00 & 55.03 & 12.20 & 24.83 & 42.08 & 50.16 & 58.17\% \\
VLM Active$^{\dagger}$
& 65.00 & 64.01 & 7.65 & 40.13 & 56.73 & 64.51 & 58.15\%\\

\midrule
\multicolumn{8}{c}{\textit{Proposed Semantic--Geometric Planning Variants}} \\
\midrule
Explicit Exploration
& 70.00 & 68.77 & 8.30 & 36.38 & 51.87 & 58.29 & 64.20\% \\
Amortized Planner
& 66.00 & 64.98 & 8.50 & 31.04 & 48.73 & 48.17 & 62.69\% \\
FUSE
& 72.00 & 70.91 & 9.32 & 34.96 & 49.52 & 56.97 & 66.61\% \\

\midrule
\multicolumn{8}{c}{\textit{Oracle-Label Reference}} \\
\midrule
Oracle-Label FUSE
& 77.00 & 75.86 & 9.98 & 39.72 & 51.70 & 59.91 & 72.36\% \\

\bottomrule
\end{tabular}
}
\vspace{1mm}
\caption{
\textbf{Active functional affordance grounding results}.
Unless otherwise stated, all methods use CRAFT-E as the affordance knowledge source.
Best results in \textbf{bold}, and second-best \underline{underlined}.
$^{\dagger}$VLM receives the oracle object label and feasible actions.
}
\label{tab:main_results}
\end{table*}

\noindent\textbf{Evidence accumulation and termination.}
At each visited viewpoint, we compute a fused target score
$r_{t,i}=s_{t,i}+d_{t,i}$ for each candidate, where $s_{t,i}$
is its SAM3 grounding score and $d_{t,i}$ is a mask-level
depth-proximity score. We track the highest fused score
$r_t=\max_i r_{t,i}$ and its associated candidate and viewpoint.
If $r_t$ improves upon the best observed score, the best grounding
is updated and the patience counter is reset; otherwise, the counter
is incremented. Exploration terminates after $K$ consecutive
non-improving observations and returns the best grounding encountered. 

\noindent\textbf{Amortized Evidence Planning.}
Although explicit functional evidence exploration provides reliable sensing decisions, repeatedly refining the 3D spatial evidence representation, rendering candidate viewpoints, and recomputing semantic--geometric uncertainty incur substantial computational cost. To reduce this overhead, we learn an amortized evidence planner that directly predicts the value of candidate sensing actions from the current observation, a candidate object hypothesis $o_i \in \mathcal{O}_q$, and the candidate action, without explicitly updating the spatial representation. When multiple candidate hypotheses are available, the hypothesis whose text embedding is most compatible with the current image is selected as the object context for the planner. The planner is trained using sampled scene transitions labeled by the measured change in SAM3 evidence, geometric uncertainty, visibility, localization, and proximity after each action. During inference, it produces values for the feasible sensing actions in a single forward pass, approximating explicit evidence planning for faster sensing decision-making. 

\noindent\textbf{FUSE: Adaptive Evidence Acquisition.} 
Explicit evidence exploration provides reliable sensing decisions but requires repeated spatial-representation refinement and uncertainty recomputation. Amortized evidence planning avoids this cost by directly predicting the values of feasible sensing actions, but its predictions may become ambiguous in unfamiliar or poorly observed states. FUSE combines both modes through confidence-gated adaptive evidence acquisition. At each step, the amortized planner predicts values $\hat{Q}_t(a;o_i)$ for the feasible actions and converts them into an action distribution whose entropy measures decision ambiguity. 
At each step, the current observation is registered with COLMAP. When entropy is below $\eta$, FUSE selects the highest-valued amortized action and defers refinement. Otherwise, it jointly refines the 3D spatial representation over all registered observations and invokes explicit semantic--geometric exploration using $\mathcal{O}_q$.
This provides a lightweight fast--deliberative strategy in which routine sensing decisions use amortized inference, while uncertain decisions trigger more costly explicit evaluation. 
Algorithm~\ref{alg:fuse} summarizes the complete procedure. FUSE begins with a sparsely initialized 3DGS representation and tracks the strongest semantic-depth-proximity score across $\mathcal{O}_q$ by balancing amortized and explicit planning until the grounding score fails to improve for $K$ steps or the sensing budget $T$ is exhausted. 
It then returns the grounding configuration and candidate label associated with the best viewpoint.

\noindent\textbf{Implementation Details.}
The spatial evidence representation is initialized from four seed RGB observations using COLMAP~\cite{schonberger2016structure} followed by incremental 3D Gaussian Splatting optimization~\cite{kerbl3Dgaussians}. Each functional query $q$ is mapped to candidate hypotheses $\mathcal{O}_q$ using neurosymbolic models, large language models, oracle object knowledge, or direct verb conditioning, and SAM3 provides pixel-level semantic masks and scores. 
For termination, each SAM3 score is summed with a mask-level depth-proximity score obtained by normalizing target depth between 0.3\,m and 1.2\,m, and exploration stops after $K$ consecutive non-improving fused scores. 
The amortized planner uses CLIP's visual backbone followed by a three-layer MLP that combines the image feature with candidate-object and action embeddings to predict action values. 
The MLP is trained to predict measured post-action evidence gain (sum of normalized post-action changes in semantic evidence, geometric certainty, visibility, localization, and proximity), and action entropy (normalized by $\log |\mathcal{A}_t|$) is computed from a unit-temperature softmax over predicted action values. 
It is trained on 416 trajectories from two scenes disjoint from evaluation using Adam, learning rate $10^{-3}$, batch size 64, and 20 epochs. 
$\eta$ was set to $0.8$ using a grid search over five randomly sampled development episodes. The stopping patience $K=5$ and sensing budget $T=30$ are fixed evaluation-protocol choices applied consistently to all active methods. 
Additional details are in the supplementary.

\section{Experimental Evaluation}
\subsubsection{Benchmark and Evaluation Setup.}

We construct a Habitat-based~\cite{habitat19iccv,szot2021habitat,puig2023habitat3} benchmark for active functional affordance grounding using procedurally assembled tabletop scenes.
The benchmark evaluates 100 independently randomized scene instances drawn from six held-out scene configurations. In each episode, a target object and 7--14 distractors are randomly sampled and placed in the workspace, and the camera is initialized from a random
viewpoint such that at least one valid target is partially occluded. The evaluation therefore varies target identity, distractor composition, object layout, occlusion, and initial viewpoint while testing on scene configurations disjoint from those used to train the amortized planner.
For episodes with multiple valid targets, a prediction is considered correct if it satisfies the success criterion for any valid target; viewpoint-quality metrics are computed with respect to the matched target.
The active sensing space is defined over a ring graph (with fixed elevation) with 36 discrete viewpoints, together with camera tilt-up and tilt-down actions. 
3DGS-based methods are initialized from four fixed seed RGB observations.

\noindent\textbf{Metrics and Evaluation Protocol.} 
Each method returns the best viewpoint encountered during exploration together with its grounding prediction. We evaluate grounding accuracy using \textit{Success Rate}, defined as the percentage of episodes in which the highest-scoring SAM3 mask at the returned viewpoint achieves an IoU greater than $0.5$ with a valid ground-truth target. We also report \textit{Average IoU} between the predicted and target masks and \textit{Average Steps}, defined as the number of active sensing actions executed after initialization. 
To measure viewpoint quality, we report four additional metrics. \textit{Target Pixel Ratio} is the min--max normalized target-pixel count at the returned viewpoint over all reachable viewpoints. \textit{Center Score} is based on the normalized distance between the target-mask centroid and the image center, with $1$ as perfect centering. \textit{Distance Score} is the inverted, min--max normalized camera--target distance over the reachable viewpoints (larger values indicate a closer view). 
Final SAM3 Score, computed by querying with the ground-truth label, is reported as a secondary quality metric. 

\begin{table}[t]
\centering
\resizebox{\columnwidth}{!}{
\begin{tabular}{lccccc}
\toprule
\textbf{Knowledge Source}
& \textbf{Success} $\uparrow$
& \textbf{IoU} $\uparrow$
& \textbf{Steps} $\downarrow$
& \textbf{Target Ratio} $\uparrow$
& \textbf{SAM3 Score} $\uparrow$ \\
\midrule
CRAFT & 53.00\% & 52.16\% & 9.99 & 36.78\% & 66.79\%\\
GPT & 62.00\% & 61.06\% & 9.35 & 33.42\% & 61.44\%\\
LLAMA & 62.00\% & 61.00\% & 9.7 & 34.63\% & 58.98\%\\
Qwen & 64.00\% & 63.05\% & 9.89 & 37.44\% & 61.61\%\\
Sonnet & 69.00\% & 67.94\% & 10.01 & 37.78\% & 66.76\%\\
Gemini & 72.00\% & 70.86\% & 9.67 & 35.51\% & 69.26\%\\
CRAFTE & 72.00\% & 70.91\% & 9.32 & 34.96\% & 66.61\%\\
Oracle-Label & 77.00\% & 75.86\% & 9.98 & 39.72\% & 72.36\%\\
\bottomrule
\end{tabular}
}
\caption{
\textbf{Robustness to different knowledge sources.}
The planner, stopping rule, and sensing budget are held fixed.
}
\label{tab:knowledge_sources}
\end{table}

\noindent\textbf{Baselines.} 
All core baselines use CRAFT-E~\cite{chen2025crafte} to generate candidate object hypotheses. \textit{Static} performs grounding from the initial observation without camera motion. \textit{Canonical View} executes one predefined motion toward the closest feasible bird's-eye viewpoint and grounds from that observation. \textit{Random Active} follows a random walk over the sensing graph, testing whether movement alone improves grounding. \textit{VLM Active} provides Gemini~\cite{comanici2025gemini} with the current observation, oracle target label, and feasible actions, allowing the VLM to control the camera until termination. \textit{Explicit Exploration} recomputes SAM3 semantic evidence and SSNR geometric uncertainty using an incrementally refined 3DGS representation at every step. \textit{Amortized Planner} selects actions using only the learned evidence planner. 
\textit{FUSE} adaptively switches between amortized planning and explicit exploration according to action entropy. Finally, Oracle-Label FUSE supplies the ground-truth object label, isolating errors due to candidate-hypothesis generation when affordance-hypothesis generation is perfect. 
We also evaluate with alternative affordance knowledge sources, including Gemini~\cite{comanici2025gemini}, Llama 3.3 70B~\cite{grattafiori2024llama}, Qwen 3.6 Plus~\cite{qwen36_plus}, GPT-5.4~\cite{openai2025gpt5system}, Claude Sonnet 4.6~\cite{ClaudeSonnet46}, and  CRAFT~\cite{chen2025craft}. 
This analysis measures whether the benefits of adaptive semantic--geometric exploration persist across different upstream functional-reasoning mechanisms.

\noindent\textbf{Active Functional Grounding.} 
Table~\ref{tab:main_results} compares FUSE with passive, active, and oracle-assisted methods on the proposed active functional affordance grounding benchmark. We observe four main trends. 
First, active perception substantially improves functional grounding under partial observability. Static grounding achieves only 42\% success, while moving once to a canonical viewpoint reduces performance to 32\%, indicating that no single predefined view consistently reveals the evidence needed for reliable grounding. Random Active improves success to 56\% using 12.20 camera actions on average, showing that additional viewpoints are beneficial even without task-directed sensing. Second, task-relevant sensing is considerably more effective than unguided exploration. Despite using more camera actions on average, Random Active remains 14 and 16 percentage points below Explicit Exploration and FUSE in success rate, respectively.
VLM Active reaches 65\% success using fewer actions and an oracle target label, demonstrating that a foundation model can perform coarse task-directed exploration. However, it still remains below FUSE despite this additional supervision. VLM Active obtains high Target Pixel Ratio and Center Score, but its Final SAM3 Score is nearly identical to Random Active (58.15\% vs.\ 58.17\%) and substantially below FUSE (66.61\%). This suggests that moving toward and centering the target does not necessarily produce the semantic and geometric evidence required for reliable grounding.
\begin{table}[t]
\centering
\resizebox{\columnwidth}{!}{
\begin{tabular}{lccccc}
\toprule
\textbf{Variant}
& \textbf{Success} $\uparrow$
& \textbf{IoU} $\uparrow$
& \textbf{Steps} $\downarrow$
& \textbf{Target Ratio} $\uparrow$
& \textbf{Center} $\uparrow$ \\
\midrule

Random Active
& 56.00 & 55.03 & 12.20 & 24.83 & 42.08\% \\

Semantic Only
& 53.00 & 52.13 & 6.98 & 29.48 & 47.60\% \\

Geometry Only
& 65.00 & 63.95 & 15.36 & 32.48 & 45.57\% \\


Amortized Only
& 66.00 & 64.98 & 8.50 & 31.04 & 48.73\% \\

FUSE
& 72.00 & 70.91 & 9.32 & 34.96 & 49.52\% \\

\bottomrule
\end{tabular}
}
\caption{
\textbf{Ablation studies} on semantic evidence, geometric uncertainty, and adaptive planning.
All variants use the same affordance knowledge source and stopping criterion.
}
\label{tab:ablation}
\end{table}
Third, the proposed planning variants provide complementary accuracy--computation trade-offs. Explicit Exploration improves the success rate from 56\% to 70\% over Random Active, while requiring substantially fewer camera actions (8.30 vs.\ 12.20). This result supports the effectiveness of the proposed semantic-- geometric acquisition surrogate. The Amortized Planner reaches 66\% success with only 8.50 actions and without repeated spatial-representation updates, showing that the sensing policy can be approximated efficiently. FUSE adaptively combines these modes and obtains the highest observed non-oracle success rate (72\%), IoU (70.91\%), and Final SAM3 Score (66.61\%).  
Finally, FUSE provides a favorable accuracy--computation trade-off relative to Explicit Exploration. On paired episodes, it preserves comparable grounding performance (72\% vs.\ 70\%; exact McNemar $p{=}0.688$) while reducing end-to-end runtime by $1.33\times$. FUSE invokes explicit exploration for only 33.41\% of decisions, requiring 2.97 3DGS refinements per episode on average; relative to Amortized Planning, this selective fallback improves success from 66\% to 72\% and IoU from 64.98\% to 70.91\%. Full paired statistics and stage-wise latency results are provided in the supplementary.

\noindent\textbf{Robustness to Affordance Knowledge Sources.} 
Table~\ref{tab:knowledge_sources} evaluates FUSE while varying only the upstream knowledge source used to generate candidate object hypotheses. FUSE remains effective across symbolic, vision--language, and large language model sources, showing that its active evidence acquisition strategy is not tied to a particular affordance reasoner. CRAFT-E achieves 72\% success, matching Gemini for the highest non-oracle success rate, while Claude Sonnet reaches 69\%; the Oracle-label variant reaches 77\%. CRAFT-E also obtains the strongest non-oracle IoU among the evaluated sources. Paired analysis shows that CRAFT-E significantly outperforms CRAFT and GPT, while its differences from Gemini, Sonnet, and the Oracle-label reference are not statistically distinguishable. 
Performance varies more substantially for weaker knowledge sources. CRAFT achieves 53\% success, while Llama, Qwen, and GPT obtain 62--64\%. Because the planner, stopping criterion, and action budget are fixed, these differences primarily reflect whether the knowledge source proposes suitable functional hypotheses. Active exploration can improve the visibility and grounding of a proposed candidate, but it cannot recover a functionally appropriate target that is absent from or poorly represented in the hypothesis set. The two stages therefore play complementary roles: the knowledge source determines \emph{what} objects may satisfy the query, whereas FUSE determines \emph{where} to acquire evidence for grounding them. 
A high Final SAM3 Score does not necessarily imply correct functional grounding, since a visible and segmentable object may correspond to an incorrect functional hypothesis.

\begin{table}[t]
\centering
\resizebox{\columnwidth}{!}{
\begin{tabular}{llccc}
\toprule
\textbf{Component} & \textbf{Variant}
& \textbf{Success}$\uparrow$
& \textbf{IoU}$\uparrow$
& \textbf{Steps}$\downarrow$ \\
\midrule

\multirow{2}{*}{Semantic Evidence}
& CLIP (Patch-level)
& 34.00 & 32.67 & 17.11\\
& SAM3 (Pixel-level)
& 70.00 & 68.84 & 8.35\\

\midrule

\multirow{4}{*}{Stopping Criterion}
& SAM3 ($K=3$)
&64.00&62.94&5.34\\
& SAM3 ($K=5$)
&70.00&68.84&8.35\\
& SAM3+Depth Proximity ($K=3$)
&69.00&67.94&6.30\\
& SAM3+Depth Proximity ($K=5$)
&72.00&70.91&9.32\\

\bottomrule
\end{tabular}
}
\caption{Implementation ablations evaluating semantic evidence granularity and grounding-based stopping criteria.}
\label{tab:implementation_ablation}
\end{table}

\noindent\textbf{Ablation Studies.} 
Table~\ref{tab:ablation} evaluates the individual contributions of semantic evidence, geometric uncertainty, and adaptive planning. 
Semantic evidence alone is insufficient for reliable active grounding:
despite using only 6.98 camera actions on average, the SAM3-driven
Semantic Only variant achieves 53\% success, below Random Active
(56\%) and substantially below FUSE (72\%). Thus, the gains cannot be
attributed to SAM3 alone; pixel-level semantic grounding must be
combined with geometric exploration and uncertainty-aware planning. 
In contrast, Geometry Only improves success to 65\% by actively revealing previously unseen regions, demonstrating that reducing geometric uncertainty provides a stronger exploration signal than semantic confidence alone. 
Combining semantic and geometric evidence improves success to 70\% under Explicit Exploration, confirming their complementary roles during viewpoint selection. FUSE further increases success to 72\% by adaptively invoking explicit reasoning when the amortized planner is uncertain.

\noindent\textbf{Effect of Granularity and Termination Conditions.}
Table~\ref{tab:implementation_ablation} evaluates key implementation choices in FUSE. 
Within the full semantic--geometric planning pipeline, replacing pixel-level SAM3 grounding with patch-level CLIP evidence reduces success from 70\% to 34\%. This result reflects the benefit of more precise semantic evidence when combined with geometric exploration; as shown in Table~\ref{tab:ablation}, SAM3-based semantic exploration alone reaches only 53\% success. 
Incorporating depth proximity into the grounding-based stopping criterion further improves success by approximately 5 percentage points over using SAM3 alone. 
The depth-proximity term reduces premature termination at distant or weakly resolved views, complementing SAM3 by favoring observations with the target at a more useful distance.

\section{Conclusion and Future Work}

We introduced \emph{Active Functional Affordance Grounding}, a new embodied perception task that requires sequentially acquiring observations to identify and spatially ground objects satisfying a functional query. To address this challenge, we proposed FUSE, an adaptive semantic--geometric evidence acquisition framework that combines explicit uncertainty-driven exploration with a learned amortized planner. 
Experiments demonstrate improved grounding success over passive,
unguided-active, and VLM-controlled baselines while reducing
computation relative to fully explicit exploration. 
Future work will validate FUSE on real robotic platforms and extend it to dynamic environments, richer manipulation tasks, and long-horizon agents that jointly learn functional reasoning, active perception, and action execution.

\textbf{Acknowledgement.} This work was supported in part by the US NSF grants IIS 2348689, IIS 2348690, IIS 2615771, and the USDA award no. 2023-69014-39716. 

\bibliography{aaai2027}

\clearpage

\section{Appendix 1}
\section{Computational Speed Up Analysis}

\begin{table}[t]
\centering
\setlength{\tabcolsep}{7pt}
\resizebox{\linewidth}{!}{
\begin{tabular}{lccc}
\toprule
Method & Success $\uparrow$ & Wall Time (h) $\downarrow$ & Speedup $\uparrow$ \\
\midrule
Amortized Planner    & 66.00 & \textbf{3.95} & 2.57$\times$ \\
Explicit Exploration & 70.00 & 10.17          & 1.00$\times$ \\
FUSE                 & \textbf{72.00} & 7.62 & 1.33$\times$ \\
\bottomrule
\end{tabular}
}
\caption{Accuracy--efficiency comparison over 100 CRAFT-E evaluation episodes. Speedup is measured relative to Explicit Exploration under the same sensing budget and stopping criterion.}
\label{tab:runtime}
\end{table}

Table~\ref{tab:runtime} compares the grounding accuracy and wall-clock cost of the three semantic--geometric planning variants over 100 evaluation episodes. Explicit Exploration requires 10.17 hours because it repeatedly refines the 3DGS representation and recomputes semantic and geometric evidence at each sensing step. The Amortized Planner reduces evaluation time to 3.95 hours, providing a $2.57\times$ speedup, but its success rate decreases from 70\% to 66\%. This indicates that direct action-value prediction substantially reduces computation, but cannot fully reproduce explicit reasoning in ambiguous or poorly observed states.

FUSE provides a more favorable accuracy--efficiency trade-off. It achieves 72\% success while requiring 7.62 hours, reducing wall-clock cost by 25.1\% relative to Explicit Exploration and yielding a $1.33\times$ speedup. Importantly, this reduction is achieved while improving success by 5 percentage points over explicit planning and 8 points over amortized planning. FUSE is slower than the fully amortized policy because uncertain decisions trigger 3DGS refinement and explicit semantic--geometric evaluation; however, these selective deliberative updates recover and exceed the accuracy of full explicit exploration without incurring its cost at every step. These results support the central design of FUSE: amortized inference efficiently handles routine sensing decisions, while explicit reasoning is reserved for states in which its additional computational cost provides the greatest benefit.

\subsection{Efficiency Analysis}
To identify the source of the computational savings, we additionally profiled FUSE over 800 sampled evaluation episodes. This profiling set is larger than the 100-episode paired benchmark and is used only for stage-wise latency and branch-frequency analysis. Calls per episode therefore need not match the camera-action averages in the main evaluation.  All GPU timings were measured with device synchronization before and after each timed operation. Calls and latency were first accumulated per episode and then averaged over the profiling set. 

Table~\ref{tab:fuse_latency} shows that FUSE routes 66.59\% of online planning decisions through the amortized branch and invokes explicit exploration for only 33.41\%. Consequently, the spatial representation is refined only $2.97\pm3.41$ times per episode on average, rather than at every online decision. This confirms that the entropy gate is actively reducing the frequency of costly spatial reasoning rather than merely adding an amortized predictor on top of an otherwise unchanged explicit pipeline. 
Amortized action prediction contributes only 0.01\% of total latency, with an average cost of 0.0027 seconds per call. The learned planner therefore introduces negligible computational overhead. In contrast, SAM3 grounding is the dominant cost, accounting for 58.94\% of total runtime, followed by view registration at 14.71\%, explicit evidence computation at 13.32\%, and 3DGS refinement at 8.16\%. Initialization, depth-based stopping, and camera execution account for the remaining 4.87\%. 
The latency breakdown also explains why FUSE can be faster than Explicit Exploration despite occasionally taking more sensing actions. FUSE does not primarily save time by shortening trajectories. Instead, it replaces most explicit planning decisions with a millisecond-scale amortized forward pass and refines the 3DGS representation only when action entropy indicates ambiguity. The resulting reduction in representation-update frequency produces the reported $1.33\times$ end-to-end speedup while retaining performance comparable to fully Explicit Exploration.
\begin{table*}[t]
\centering
\small
\setlength{\tabcolsep}{3.5pt}
\caption{FUSE latency and adaptive-planning statistics over 800 evaluation
episodes with different knowledge sources.}
\label{tab:fuse_latency}
\begin{tabular}{lrrrr}
\toprule
\textbf{Component} &
\textbf{Calls/Ep.} &
\textbf{Time/Call (s)} &
\textbf{Time/Ep. (s)} &
\textbf{Share (\%)} \\
\midrule
Initialization                     & 1.00   & 13.2707 & 13.2707 & 4.56 \\
SAM3 grounding                     & 9.74   & 17.6225 & 171.6019 & 58.94 \\
Depth and stopping                 & 9.74   & 0.0061  & 0.0598 & 0.02 \\
Amortized planning                 & 8.74   & 0.0027  & 0.0233 & 0.01 \\
View registration                  & 9.74   & 4.3974  & 42.8280 & 14.71 \\
3DGS refinement                    & 2.97   & 8.0051  & 23.7851 & 8.16 \\
Explicit evidence computation      & 9.74   & 3.9826  & 38.7810 & 13.32 \\
Camera execution                   & 18.48  & 0.0444  & 0.8201 & 0.28 \\
\midrule
\textbf{Total}                     & --   & --  & \textbf{291.1699} & 100 \\
\midrule
Amortized branch rate              & \multicolumn{4}{c}{66.59\%} \\
Explicit branch rate               & \multicolumn{4}{c}{33.41\%} \\
3DGS refinements per episode       & \multicolumn{4}{c}{2.97 $\pm$ 3.41} \\
\bottomrule
\end{tabular}
\end{table*}

\section{Paired Statistical and Runtime Analysis}
\label{sec:supp_stats_runtime}

\subsection{Paired Statistical Protocol}

All planning variants were evaluated on the same 100 episode definitions, enabling paired comparisons at the episode level. For binary grounding success, we report the exact McNemar test, the paired success-rate difference, a 95\% percentile-bootstrap confidence interval computed from 10,000 paired resamples, and the number of episodes uniquely solved by each method. 
For continuous metrics, we report the paired mean difference and its 95\% paired-bootstrap confidence interval. We additionally use the two-sided Wilcoxon signed-rank test and apply Holm correction within each metric across the two comparisons:

FUSE versus Explicit Exploration and FUSE versus Amortized

Planning. The bootstrap intervals quantify effect size uncertainty,

whereas the Wilcoxon tests assess paired distributional differences. We do not interpret an effect as statistically reliable solely from an uncorrected test.

The logged \texttt{steps\_T} values count online sensing actions. The four initialization actions required by 3DGS-based methods are included in the total camera-action counts reported in the main paper. Adding the same four initialization actions to FUSE and Explicit Exploration does not alter their paired action difference.

\subsection{Paired Grounding Success}

\begin{table}[t]
\centering
\small
\setlength{\tabcolsep}{4.5pt}
\caption{Paired grounding-success analysis over the shared 100 evaluation episodes. ``FUSE only'' and ``Other only'' count discordant episodes uniquely solved by each method. Confidence intervals are computed from 10,000 paired bootstrap resamples.}
\label{tab:paired_success}
\resizebox{\linewidth}{!}{
\begin{tabular}{lccccccc}
\toprule
\textbf{Comparison} &
\textbf{FUSE} &
\textbf{Other} &
\textbf{$\Delta$} &
\textbf{95\% CI} &
\textbf{FUSE only} &
\textbf{Other only} &
\textbf{McNemar $p$} \\
\midrule
Explicit Exploration
& 72\% & 70\% & $+2$ pp
& $[-3,\,7]$ pp
& 4 & 2 & 0.688 \\
Amortized Planning
& 72\% & 66\% & $+6$ pp
& $[0,\,12]$ pp
& 8 & 2 & 0.109 \\
\bottomrule
\end{tabular}
}
\end{table}

Table~\ref{tab:paired_success} shows that FUSE preserves comparable grounding success relative to fully Explicit Exploration. FUSE solves four episodes that Explicit Exploration misses, while Explicit Exploration uniquely solves two, yielding a two-point paired difference whose confidence interval includes zero ($p=0.688$). Thus, the current evaluation does not establish a statistically distinguishable success advantage over fully explicit reasoning. Instead, the primary benefit of FUSE in this comparison is computational: it retains comparable grounding performance while avoiding repeated spatial-representation refinement.
Relative to Amortized Planning, FUSE improves success by six percentage points and uniquely solves eight episodes, compared with two episodes uniquely solved by the amortized variant. The exact McNemar test is not significant at the conventional 0.05 level ($p=0.109$), reflecting the small number of discordant outcomes. Nevertheless, the four-to-one asymmetry in uniquely solved episodes is consistent with the intended role of explicit fallback: recovering difficult cases in which amortized action prediction alone is insufficient.

\subsection{Paired Continuous-Metric Analysis}

\begin{table*}[t]
\centering
\small
\setlength{\tabcolsep}{4.2pt}
\caption{Paired continuous-metric comparisons over 100 episodes. Differences are computed as FUSE minus the comparison method. For Online Actions, positive differences indicate that FUSE uses more sensing actions; for all other metrics, positive differences favor FUSE. The final column reports Holm-adjusted two-sided
Wilcoxon signed-rank $p$-values.}
\label{tab:paired_continuous}
\begin{tabular}{llrrrrr}
\toprule
\textbf{Comparison} &
\textbf{Metric} &
\textbf{FUSE} &
\textbf{Other} &
\textbf{Paired $\Delta$} &
\textbf{95\% CI} &
\textbf{Holm $p$} \\
\midrule
\multirow{7}{*}{Explicit Exploration}
& IoU
& 0.709 & 0.688 & $+0.021$
& $[-0.026,\,0.071]$ & 0.809 \\
& Online Actions
& 9.32 & 8.30 & $+1.02$
& $[0.31,\,1.79]$ & \textbf{0.012} \\
& Target Pixel Ratio
& 0.350 & 0.364 & $-0.014$
& $[-0.065,\,0.035]$ & 0.629 \\
& Target Pixel Ratio$_{90}$
& 0.561 & 0.547 & $+0.014$
& $[-0.050,\,0.078]$ & 0.407 \\
& Best Target Score
& 0.586 & 0.564 & $+0.022$
& $[-0.011,\,0.056]$ & 0.487 \\
& Center Score
& 0.495 & 0.519 & $-0.024$
& $[-0.067,\,0.021]$ & 0.628 \\
& Geometric Score
& 0.570 & 0.583 & $-0.013$
& $[-0.073,\,0.045]$ & 0.854 \\
\midrule
\multirow{7}{*}{Amortized Planning}
& IoU
& 0.709 & 0.650 & $+0.059$
& $[0.001,\,0.119]$ & 0.809 \\
& Online Actions
& 9.32 & 8.50 & $+0.82$
& $[-0.34,\,1.90]$ & \textbf{0.020} \\
& Target Pixel Ratio
& 0.350 & 0.310 & $+0.039$
& $[-0.005,\,0.086]$ & 0.076 \\
& Target Pixel Ratio$_{90}$
& 0.561 & 0.479 & $+0.082$
& $[0.023,\,0.143]$ & \textbf{0.032} \\
& Best Target Score
& 0.586 & 0.556 & $+0.030$
& $[-0.020,\,0.081]$ & 0.167 \\
& Center Score
& 0.495 & 0.487 & $+0.008$
& $[-0.041,\,0.057]$ & 0.628 \\
& Geometric Score
& 0.570 & 0.482 & $+0.088$
& $[0.026,\,0.150]$ & \textbf{0.004} \\
\bottomrule
\end{tabular}
\end{table*}

The continuous metrics reinforce the interpretation of the success results. Relative to Explicit Exploration, FUSE produces similar IoU, target exposure, centering, and geometric quality: all corresponding confidence intervals include zero, and none of the adjusted tests is significant. FUSE does use 1.02 additional online actions on average ($p_{\mathrm{Holm}}=0.012$). This does not contradict the measured runtime advantage because the additional decisions are predominantly handled through inexpensive amortized inference rather than costly 3DGS refinement. FUSE therefore trades a small increase in sensing actions for substantially lower representation-update cost. 
Relative to Amortized Planning, FUSE improves mean IoU by 0.059, with a paired-bootstrap interval of $[0.001,0.119]$. The Holm-adjusted Wilcoxon result is not significant, so we treat this as suggestive rather than conclusive evidence of an IoU improvement. More diagnostic viewpoint metrics show clearer gains. FUSE improves the robust target exposure measure, Target Pixel Ratio$_{90}$, by 0.082 ($p_{\mathrm{Holm}}=0.032$), and improves the geometric score by 0.088 ($p_{\mathrm{Holm}}=0.004$). These results indicate that the explicit fallback helps FUSE acquire viewpoints with stronger target exposure and geometric evidence than the amortized planner alone.
FUSE uses 0.82 additional online actions relative to Amortized Planning. Although the Wilcoxon test identifies a distributional difference, the paired-bootstrap interval for the mean difference includes zero. The combined evidence therefore suggests that FUSE occasionally explores longer, but uses those additional observations to improve geometric coverage and recover episodes that amortized planning alone fails to ground. 
The logged \texttt{best\_nav\_score} was constant at $-1$ for all methods and episodes and was therefore excluded from Table~\ref{tab:paired_continuous} and inferential analysis.

\subsubsection{Paired Analysis of Functional Knowledge Sources}
\label{sec:supp_knowledge_paired} 
We evaluate each knowledge source on the same 100 episodes while holding the FUSE planner, grounding model, stopping criterion, and action budget fixed. Table~\ref{tab:knowledge_source_paired} reports paired success differences relative to CRAFT-E. Confidence intervals are obtained from 10,000 paired bootstrap resamples, and exact McNemar $p$-values are Holm-adjusted across all knowledge-source comparisons.
\begin{table}[t]
\centering
\setlength{\tabcolsep}{4pt}
\caption{Paired success analysis over the shared 100 episodes. CRAFT-E is the reference. $\Delta$ denotes CRAFT-E minus the comparison source, and unique wins are reported as CRAFT-E/comparison.}
\label{tab:knowledge_source_paired}
\resizebox{\linewidth}{!}{
\begin{tabular}{lrrrrr}
\toprule
\textbf{Source} &
\textbf{Success} &
\textbf{$\Delta$} &
\textbf{95\% CI} &
\textbf{Unique wins} &
\textbf{Adj.\ $p$} \\
\midrule
CRAFT         & 53\% & $+19$ pp & $[10,28]$ & 23/4 & \textbf{0.002} \\
GPT           & 62\% & $+10$ pp & $[4,17]$  & 11/1 & \textbf{0.038} \\
Llama         & 62\% & $+10$ pp & $[3,18]$  & 13/3 & 0.106 \\
Qwen          & 64\% & $+8$ pp  & $[1,15]$  & 11/3 & 0.229 \\
Claude Sonnet & 69\% & $+3$ pp  & $[-4,10]$ & 8/5  & 1.000 \\
Gemini        & 72\% & $0$ pp   & $[-6,6]$  & 5/5  & 1.000 \\
Oracle label  & 77\% & $-5$ pp  & $[-11,1]$ & 2/7  & 0.539 \\
\bottomrule
\end{tabular}
}
\end{table}
CRAFT-E significantly improves success over CRAFT ($+19$ pp, adjusted $p{=}0.002$) and GPT ($+10$ pp, adjusted $p{=}0.038$). Its observed improvements over Llama and Qwen do not remain significant after correction, while its performance is statistically indistinguishable from Claude Sonnet, Gemini, and the Oracle-label reference. In particular, CRAFT-E and Gemini both achieve 72\% success, whereas the Oracle reaches 77\%. Online sensing actions do not differ significantly across knowledge sources, indicating that the success differences primarily reflect candidate-hypothesis quality rather than unequal exploration budgets.

\section{Implementation Details}
\label{sec:supp_implementation}

\subsection{Environment and Evaluation Episodes}

Experiments are implemented in Habitat-Sim v0.3.3 and Habitat-Lab v0.3.3, using the Habitat-compatible AI2-THOR (iTHOR) scene dataset. The benchmark uses six evaluation scene configurations that are disjoint from the two configurations used to train the amortized planner. We evaluate all methods on the same fixed set of 100 episode definitions. Each definition stores the scene configuration, functional query, valid target objects, distractor objects, object poses, initial camera pose, and random seed.  Each episode contains one sampled target and 7--14 sampled distractors. Object layouts, including any additional object placements, are fixed within each scene configuration. 

\subsection{Camera and Action Space}

For each scene configuration, we construct a scene-specific elliptical camera
trajectory in the horizontal plane. The ellipse is parameterized by a
scene-dependent center and major and minor axes, selected to provide
collision-free views of the workspace. Each trajectory is uniformly discretized
into 36 camera positions at $10^\circ$ angular intervals. Camera height is scene-specific and ranges from 1.4 to 1.6~m.
At each position, the nominal camera orientation faces the ellipse center in
the horizontal plane. The scene-specific trajectory parameters are specified
in the accompanying code submission.

Each trajectory position is paired with three discrete downward pitch angles,
$\{15^\circ, 30^\circ, 45^\circ\}$, yielding a discrete viewpoint state space.
An action is represented by a joint displacement
$(\Delta p, \Delta r)$, where $\Delta p \in \{-2,-1,0,+1,+2\}$ denotes motion
along the elliptical trajectory in units of viewpoint nodes, and
$\Delta r \in \{-1,0,+1\}$ denotes a change in pitch level. The available
actions comprise a no-op action, one- and two-node horizontal motions,
one-level pitch adjustments, and their horizontal--pitch combinations. Actions
that would produce an invalid pitch level are masked.

For explicit exploration, horizontal action components are associated with the
corresponding left or right image region, whereas pitch-up and pitch-down
components are associated with the upper and lower image regions, respectively.
For coupled actions, we combine the evidence associated with their horizontal
and vertical components.

\subsection{3D Gaussian Splatting}

For each episode, all 3DGS-based methods use the same four initialization observations and camera motions. These observations are excluded from the online budget $T$, ineligible as returned output viewpoints, and are included in the reported end-to-end runtime. 
Camera poses and the initial sparse reconstruction are obtained using COLMAP
v3.13.0. For initialization, we extract features from the initialization
observations, perform exhaustive matching, and run COLMAP's incremental mapper.
During mapping, focal length, principal point, and extra camera parameters are
held fixed.

For each subsequent observation, we extract features and apply sequential
matching with an overlap of 10. We enable guided matching, set the maximum
number of matches to 32,768, use a SIFT ratio threshold of 0.9, and set the
maximum SIFT descriptor distance to 0.8. The observation is registered into
the existing sparse model, followed by point triangulation and bundle
adjustment. The camera intrinsics remain fixed during incremental bundle
adjustment; all other COLMAP settings use the defaults of v3.13.0.

We use the official implementation of \emph{3D Gaussian Splatting for
Real-Time Radiance Field Rendering}~\cite{kerbl3Dgaussians}, commit
\texttt{54c035f7834b}. Initial optimization uses 1000 iterations. Incremental updates use 400 iterations after each explicit-planning observation. The main 3DGS settings are:

\begin{itemize}
    \item image resolution: $640 \times 480$;
    \item spherical-harmonic degree: $3$;
    \item position learning rate: $1.6 \times 10^{-4}$ to
    $1.6 \times 10^{-6}$;
    \item feature learning rate: $2.5 \times 10^{-3}$;
    \item opacity learning rate: $2.5 \times 10^{-2}$;
    \item scale learning rate: $5.0 \times 10^{-3}$;
    \item rotation learning rate: $1.0 \times 10^{-3}$;
    \item densification interval: $100$ iterations;
    \item densification start/end iterations: $500$ / $15{,}000$;
    \item opacity-reset interval: $3{,}000$ iterations.
\end{itemize}

FUSE does not update the 3DGS representation on amortized-planning steps. When explicit planning is invoked, optimization resumes from the most recently updated representation.

\subsection{Semantic and Geometric Evidence}

We use SAM3 with the \texttt{sam3.pt} checkpoint to obtain dense semantic
evidence from $640 \times 480$ RGB observations. Given the candidate object
labels for a functional query, SAM3 is applied independently to each label.
For candidate label $i$, let $m_{t,i,j}$ and $s_{t,i,j}$ denote the $j$-th
predicted mask and its confidence score at time $t$, respectively. We construct
the semantic evidence map by assigning each mask its confidence score and
taking the pixelwise maximum over all candidate labels and predicted masks:
\[
S_t(u)
=
\max_{i,j}
s_{t,i,j}\mathbb{1}[u \in m_{t,i,j}].
\]
This preserves the strongest candidate-relevant semantic evidence at each
pixel for subsequent viewpoint selection.

To measure geometric uncertainty, we compare the current RGB observation with
the corresponding rendering from the current 3DGS representation. Both images
are converted to grayscale, and gradient magnitudes are computed using
$3 \times 3$ Sobel filters. Let $I_t$ and $R_t$ denote the observed and
rendered images, respectively, and let $\mathcal{N}_{11}(u)$ denote an
$11 \times 11$ local window centered at pixel $u$. We define
\[
A_t(u)
=
\sum_{v \in \mathcal{N}_{11}(u)}
\lVert \nabla I_t(v) \rVert_2^2.
\]
\[
B_t(u)
=
\sum_{v \in \mathcal{N}_{11}(u)}
\left(
\lVert \nabla I_t(v) \rVert_2 -
\lVert \nabla R_t(v) \rVert_2
\right)^2.
\]
The local structural signal-to-noise ratio is then
\[
\operatorname{SSNR}_t(u)
=
10\log_{10}
\left(
\frac{A_t(u)+\epsilon}{B_t(u)+\epsilon}
\right),
\qquad
\epsilon=10^{-8}.
\]
We use reflection padding at image boundaries. We clip SSNR values to
$[-30,30]$ dB and linearly normalize them to $[0,1]$. Geometric uncertainty is
defined as
\[
U_t^{\mathrm{geo}}(u)
=
\alpha
\left[
1 -
\frac{
\operatorname{clip}(\operatorname{SSNR}_t(u),-30,30)+30
}{60}
\right],
\]
where $\alpha=0.8$ in all experiments. If either the observed image or its
corresponding rendering is unavailable, we set the SSNR map to zero.

We fuse semantic evidence and geometric uncertainty as
\[
E_t(u)
=
w_s S_t(u)
+
w_g U_t^{\mathrm{geo}}(u),
\]
where $w_s=w_g=1$ in all experiments.

\subsection{Depth-Proximity Stopping Rule}

For each candidate region $m_{t,i}$, we compute the mean of valid positive
depth values within the region using the Habitat depth observation. Depth
values are converted to meters using a scale factor of 1,000. A candidate
region is considered valid only if it contains at least 500 pixels. Let
$\bar{z}_{t,i}$ denote its mean depth. The depth-proximity score is
\[
d_{t,i}
=
1 -
\operatorname{clip}
\left(
\frac{\bar{z}_{t,i}-0.3}{1.2-0.3},
0, 1
\right).
\]
If a candidate region contains no valid depth value, we set $d_{t,i}=0$.

We combine semantic confidence and depth proximity as
\[
r_{t,i}
=
\lambda_s s_{t,i}
+
\lambda_d d_{t,i},
\]
where $\lambda_s=0.7$ and $\lambda_d=0.3$ in all experiments. We retain the
viewpoint associated with the largest grounding score observed so far.
Exploration terminates after $K=5$ consecutive observations that do not improve the best observed score, or after $T=30$ online actions.

\subsection{Amortized Planner Architecture}

The planner uses the frozen CLIP ViT-B/16 image and text encoders. The RGB images are resized to $224 \times 224$ and processed using the standard CLIP normalization. The image and text encoders each produce 512-dimensional features.  At each step, the object context is selected as the candidate label whose text embedding has the highest cosine similarity with the current image embedding. Actions are represented using 32-dimensional learned embeddings.

The image, object, and action features are directly concatenated. They are processed by a three-layer MLP with two hidden layers:
\[
1056 \rightarrow 512 \rightarrow 512 \rightarrow 1.
\]
The hidden layers use ReLU activations. CLIP remains frozen during training.

\subsection{Planner Training Targets}

We collect 416 trajectories from scene configurations disjoint from evaluation,
comprising 3479 transitions. Each trajectory is initialized from a randomly
sampled viewpoint and follows a random rollout of 5--13 actions sampled
uniformly from the discrete action set. We randomly split the resulting
transitions into 90\% training and 10\% validation sets.

For a transition $(x_t,a_t,x_{t+1})$, we define the utility target as
\[
y_t
=
w_{\mathrm{sem}}\Delta s_t
+
w_{\mathrm{unc}}\Delta u_t
+
w_{\mathrm{iou}}\Delta q_t
+
w_{\mathrm{vis}}\Delta v_t
+
w_{\mathrm{geo}}\Delta g_t.
\]
Individual terms are defined as
\[
\Delta s_t = s_{t+1}-s_t,
\qquad
\Delta u_t = u_t-u_{t+1}.
\]
\[
\Delta q_t = q_{t+1}-q_t,
\qquad
\Delta v_t = v_{t+1}-v_t.
\]
\[
\Delta g_t = g_{t+1}-g_t.
\]
Here, $s_t$ denotes the maximum SAM3 prompt confidence, $u_t$ denotes the mean
3DGS uncertainty over the nonzero semantic-evidence region, $q_t$ denotes the
IoU between the highest-confidence SAM3 region and the ground-truth target
mask, $v_t$ denotes target visibility normalized by the scene-specific maximum
target pixel count, and $g_t$ denotes normalized inverse camera-to-target
distance. We use $w_{\mathrm{sem}}=1.0$, $w_{\mathrm{unc}}=0.5$,
$w_{\mathrm{iou}}=1.0$, $w_{\mathrm{vis}}=0.2$, and
$w_{\mathrm{geo}}=1.0$. Targets are clipped to $[-1,1]$.

The planner is trained using mean-squared error and Adam for 20 epochs, with
a learning rate $10^{-3}$ and batch size 64. We use no weight decay or gradient
clipping. We save the model after the final epoch; validation loss is monitored
but is not used for checkpoint selection.

\subsection{Confidence Gate}

For predicted action values $\hat Q_t(a)$, action probabilities are computed using a softmax with temperature $\tau=0.1$. We use normalized entropy:
\[
H_t = -\frac{1}{\log|\mathcal A_t|} \sum_{a\in\mathcal A_t} p_t(a)\log p_t(a).
\]

FUSE uses amortized planning when $H_t\leq0.8$ and explicit exploration otherwise. The threshold was selected from $\{0.6, 0.7, 0.8, 0.9\}$ using 5 validation episodes from scene configurations disjoint from the evaluation set.  Across the evaluation set, 66.59\% of decisions use the amortized branch and 33.41\% use explicit exploration. FUSE performs 2.97 3DGS updates per episode on average.

\subsection{Baseline Settings}

All methods use the same fixed episode definitions, candidate hypotheses, SAM3 model, stopping rule, and online action budget unless otherwise stated.

\paragraph{Random Active.}
Actions are sampled uniformly from the feasible action set. We use a base random
seed of $2$ and set the episode seed to $2+e$ for episode $e$, which controls
random action sampling and environment initialization. Results are reported over
the same fixed set of evaluation episodes used for all methods.

\paragraph{Canonical View.}
Grounding is performed from a fixed top-down view located above and oriented toward the scene center.

\paragraph{VLM Active.}
We use Gemini-3.1-Flash-Lite with temperature 0 for active viewpoint
selection. At each step, the model receives the current RGB observation and
an aligned depth visualization, the candidate object labels associated with
the target verb, the current discrete loop position and pitch, the feasible
actions, and the six most recent action-state records. The camera moves on a
discrete elliptical loop around the scene, with horizontal actions
\texttt{\{stay, move\_left\_10, move\_right\_10, move\_left\_20,
move\_right\_20\}} and pitch actions
\texttt{\{pitch\_keep, pitch\_up, pitch\_down\}}. The available downward
pitch angles are \(15^\circ\), \(30^\circ\), and \(45^\circ\).

The model returns a schema-constrained JSON decision specifying a
stop/continue status, one horizontal action, one pitch action, and predicted
target visibility, centering, and reachability. It is instructed to stop only
when a candidate object is visible, centered, and sufficiently close for
manipulation; otherwise, it continues searching or refining the view. An
episode terminates when the model outputs \texttt{stop} while predicting that
the target is visible, or after 30 decision steps. Grounding is then performed
on the final RGB observation. Invalid or failed model responses trigger a
conservative no-motion \texttt{continue} action. API latency is excluded from
runtime.

\paragraph{Oracle-Label FUSE.}
Only the generated candidate set is replaced by the ground-truth object label. No oracle masks, poses, actions, or stopping signals are used.

\subsection{Metric Computation}

For each target object in each scene, we precompute statistics over the
candidate camera viewpoints, including the maximum visible target-pixel count
\(P_{\max}\) and the minimum and maximum camera-to-target distances,
\(D_{\min}\) and \(D_{\max}\). These statistics are used only for metric
normalization.

At the returned viewpoint, i.e., the highest-scoring viewpoint encountered during exploration, SAM3 is prompted with the candidate object set associated with the target verb. We select its highest-scoring predicted
instance and compute IoU between its mask and the ground-truth semantic mask
of the target object. Success is defined as \(\mathrm{IoU} \geq 0.5\); if
SAM3 returns no instance mask, IoU is set to zero.

Target Pixel Ratio and Distance Score are computed as
\[
\mathrm{PixelRatio} =
\operatorname{clip}\left(
\frac{P-P_{\min}}{P_{\max}-P_{\min}+\epsilon}, 0, 1
\right),
\]
\[
\mathrm{DistanceScore} =
\operatorname{clip}\left(
\frac{D_{\max}-D}{D_{\max}-D_{\min}+\epsilon}, 0, 1
\right).
\]
where \(P\) is the target-mask pixel count and \(D\) is the Euclidean distance
between the returned camera and target positions. Center Score is one minus the
target-mask centroid's distance to the image center, normalized by the
center-to-corner distance.

Finally, Final SAM3 Score is the confidence of the highest-scoring SAM3
instance queried with the ground-truth object label. It is a secondary
view-quality metric; Success and IoU use ground-truth semantic masks.

\subsection{Runtime Protocol}

Runtime is measured per episode and includes initialization-view acquisition,
COLMAP initialization and incremental view registration, initial and incremental 3D Gaussian Splatting (3DGS) optimization, SAM3 inference, evidence computation, action selection, camera-pose execution, and final-view evaluation. For methods without 3D reconstruction, runtime includes their
corresponding view-selection and grounding operations.

Runtime excludes environment construction and scene loading, model loading, and external hosted-API latency. Episodes are executed sequentially on a single worker, and we report end-to-end runtime over the same 100 evaluation episodes.

\subsection{Hardware and Software}

Experiments are conducted on a workstation with:
\begin{itemize}
    \item CPU: AMD Ryzen Threadripper PRO 5995WX (64 cores);
    \item system memory: 503~GB;
    \item GPU: four NVIDIA RTX A5500 GPUs, each with 24~GB memory;
    \item operating system: Ubuntu 24.04.4 LTS;
    \item NVIDIA driver: 580.173.02;
    \item simulation stack: Habitat-Sim 0.3.3 and Habitat-Lab 0.3.3
    (Python 3.9.23, PyTorch 2.5.1+cu121, CUDA 12.1);
    \item grounding stack: \texttt{facebook/sam3}
    (Python 3.12.13, PyTorch 2.10.0+cu128, CUDA 12.8);
    \item reconstruction stack: COLMAP 3.13.0
    (Python 3.8.20, PyTorch 2.1.2+cu118, CUDA 11.8).
\end{itemize}

The amortized planner requires approximately 0.10 hours (6 minutes) to train and uses 0.46~GB peak GPU memory. 
\end{document}